%% file: maincorl.tex
\documentclass{article}

\usepackage[final]{corl_2017} 

\usepackage{hyperref, url, amsmath, amssymb, booktabs, graphicx}
\graphicspath{{images/}}
\usepackage[export]{adjustbox}

\definecolor{blue}{RGB}{69,150,236}
\definecolor{orange}{RGB}{247,194,68}
\definecolor{plum}{RGB}{144,54,170}
\definecolor{green}{RGB}{139,195,74}

\title{Semi-Supervised Haptic Material Recognition for Robots using Generative Adversarial Networks}

\author{
  Zackory Erickson, Sonia Chernova, and Charles C. Kemp \\
  Institute for Robotics and Intelligent Machines \\
  Georgia Institute of Technology, 
  United States \\
  \texttt{zackory@gatech.edu}, \texttt{chernova@cc.gatech.edu}, \texttt{charlie.kemp@bme.gatech.edu} \\
}

\begin{document}
\maketitle

\vspace{-1em}
\begin{abstract}
Material recognition enables robots to incorporate knowledge of material properties into their interactions with everyday objects. For example, material recognition opens up opportunities for clearer communication with a robot, such as ``bring me the metal coffee mug", and recognizing plastic versus metal is crucial when using a microwave or oven. However, collecting labeled training data with a robot is often more difficult than unlabeled data. We present a semi-supervised learning approach for material recognition that uses generative adversarial networks (GANs) with haptic features such as force, temperature, and vibration. Our approach achieves state-of-the-art results and enables a robot to estimate the material class of household objects with $\sim$90\% accuracy when 92\% of the training data are unlabeled. We explore how well this approach can recognize the material of new objects and we discuss challenges facing generalization. To motivate learning from unlabeled training data, we also compare results against several common supervised learning classifiers. In addition, we have released the dataset used for this work which consists of time-series haptic measurements from a robot that conducted thousands of interactions with 72 household objects.
\end{abstract}



\section{Introduction}
\label{sec:intro}

The ability to recognize materials plays an important role in how we interpret our surroundings and interact with the environment. For example, touch-based material recognition can enable us to infer objects when reaching into occluded spaces. If we reach into a purse and feel metal, it is likely either keys or coins. In addition, recognizing materials allows us to decide if an object can be heated safely in a microwave. Yet, material recognition of real-world objects remains challenging for robots. 

In comparison to material recognition, the field of computer vision has seen substantial improvements recently in part due to the availability of massive image and video datasets. Given this, one approach to learn material recognition is to collect a large dataset of labeled training data that covers most of the variation a robot will encounter in the real world. However, robots are bound to interact with new objects that differ significantly from their original training set. Oftentimes, there is also a high cost associated with collecting labeled data on a robot.

\begin{figure}
\centering
\includegraphics[width=0.60\textwidth, trim={6cm 0cm 0cm 0cm}, clip]{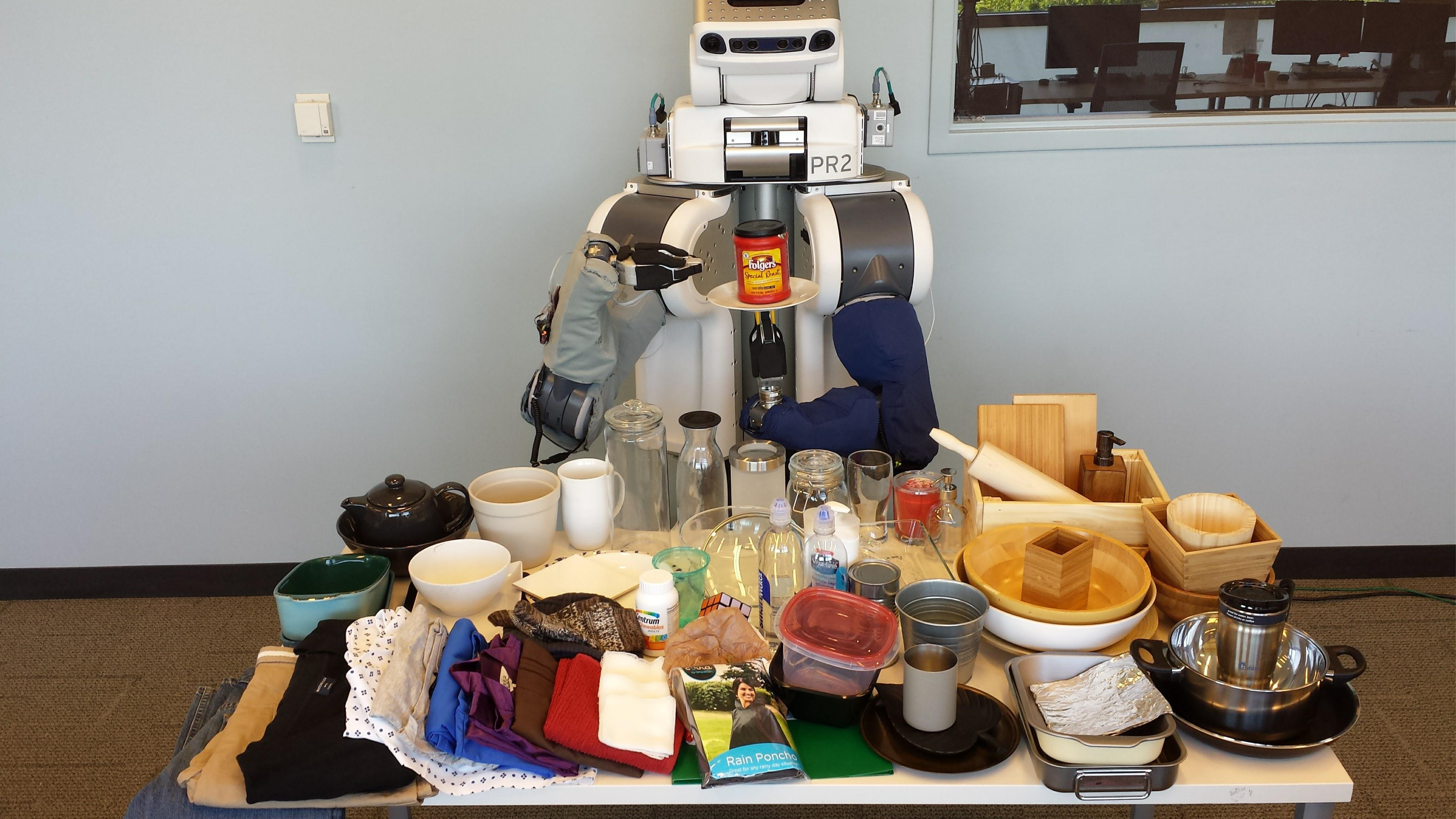}
\caption{\label{fig:pr2objects}A PR2 robot along with the 72 objects that it interacted with.}
\vspace{-1em}
\end{figure}

In this work, we present a semi-supervised learning approach with generative adversarial networks (GANs) that enables a robot to learn from unlabeled tactile sensory data from interactions with everyday objects\footnote{All code can be found at: \url{https://github.com/healthcare-robotics/mr-gan}}.
By leveraging unlabeled sensor data that are more abundant in unstructured environments, we mitigate the need for massive labeled training sets. For example, our robot can classify the materials of household objects with $\sim$90\% accuracy, when 92\% of the training data are unlabeled. This generative model can recognize an object's material with 90\% accuracy given just half a second of contact. We also explore how well this approach can recognize the materials of new objects. We note that further work is necessary, and we discuss some current challenges facing generalization. Finally, we are releasing a dataset\footnote{Dataset can be downloaded at: \url{http://healthcare-robotics.com/mr-gan}} of time-series tactile measurements from a robot that conducted thousands of interactions with 72 household objects, as seen in Figure~\ref{fig:pr2objects}. 

\section{Related Work}
\label{sec:relatedwork}

\subsection{Haptic Perception}

\citet{bell15minc} introduced the Materials in Context Database (MINC) for material classification of everyday images. However, there are many scenarios in which it is challenging to visually determine an object's material. For example, objects can be colored to alter their material appearance, like non-metal objects with metallic paint, or an object could be visually occluded or surrounded by clutter. 

\citet{Bhattacharjee2015MaterialRF} presented how a 1-DoF robot can perform supervised material recognition using active temperature sensing over a short duration of contact. Their robot distinguished between 11 materials by interacting with flat blocks of each raw material. Using an SVM on 500 labeled interactions per material, they achieved a material classification accuracy of 84\% with 0.5 seconds of contact and 98\% with 1.5 seconds of contact. However, note that this accuracy may not transfer to real-world objects with curved or uneven surfaces~\citep{bhattacharjee2016data}. In our work, we also used active temperature sensors, and we compare GAN performance to more classical approaches such as SVMs.

\citet{kerr2013material} also explored thermal sensing for material classification. They used a BioTac sensor to interact with 6 materials and collected 15 trials per material. Using a neural network, they achieved a classification accuracy of 73\% given 20 seconds of contact per trial. \citet{decherchi2011tactile} used force sensing for material classification. They compared neural networks, regularized least squares, and SVMs. Their best model achieved a classification accuracy of 89.5\% using 300 training examples across four materials. \citet{Chu2013UsingRE} used a PR2 with two BioTac sensors to estimate haptic adjectives (e.g. soft, slippery, or fuzzy) of 51 everyday objects. \citet{Sinapov2014632} used a robot with Barrett WAM arms for object recognition of 100 objects spanning 20 categories.

While many modern learning techniques have been relatively underexplored in haptic perception, there are a few works that have employed deep learning. \citet{bell15minc} used visual features from the MINC dataset to fine-tune a convolutional neural network (CNN) model for material classification. \citet{gao2016deep} coupled this visual CNN model with a CNN for BioTac haptics data to estimate haptic adjectives. \citet{yuan2017shape} showed how long short-term memory and CNN networks can learn the hardness of an object given visual tactile features from a GelSight sensor. Note that all of these works rely on visual features and network models pretrained on large, labeled image datasets, such as VGG-16~\citep{simonyan2014very}. This is due in part to the complexity of collecting labeled haptics data with a robot.

\subsection{Semi-Supervised Learning}

Several past works have considered semi-supervised learning for the task of image classification. This can be seen in ladder networks by~\citet{NIPS2015_5947} and deep generative networks by~\citet{NIPS2014_5352}. We leverage a deep neural network architecture similar to these works, but we also use adversarial training in order to improve material classification results.

There is little research in robotics on semi-supervised learning for haptic perception. One exception is work by~\citet{Luo2015} who explored semi-supervised learning for object recognition via ensemble manifold regularization. They used a Phantom haptic device to interact with 12 objects. However, unlike our work, they focused on object recognition and interacted with each object only five times.

\begin{figure}
\centering
\includegraphics[width=0.30\textwidth, trim={0cm 6cm 0cm 2cm}, clip]{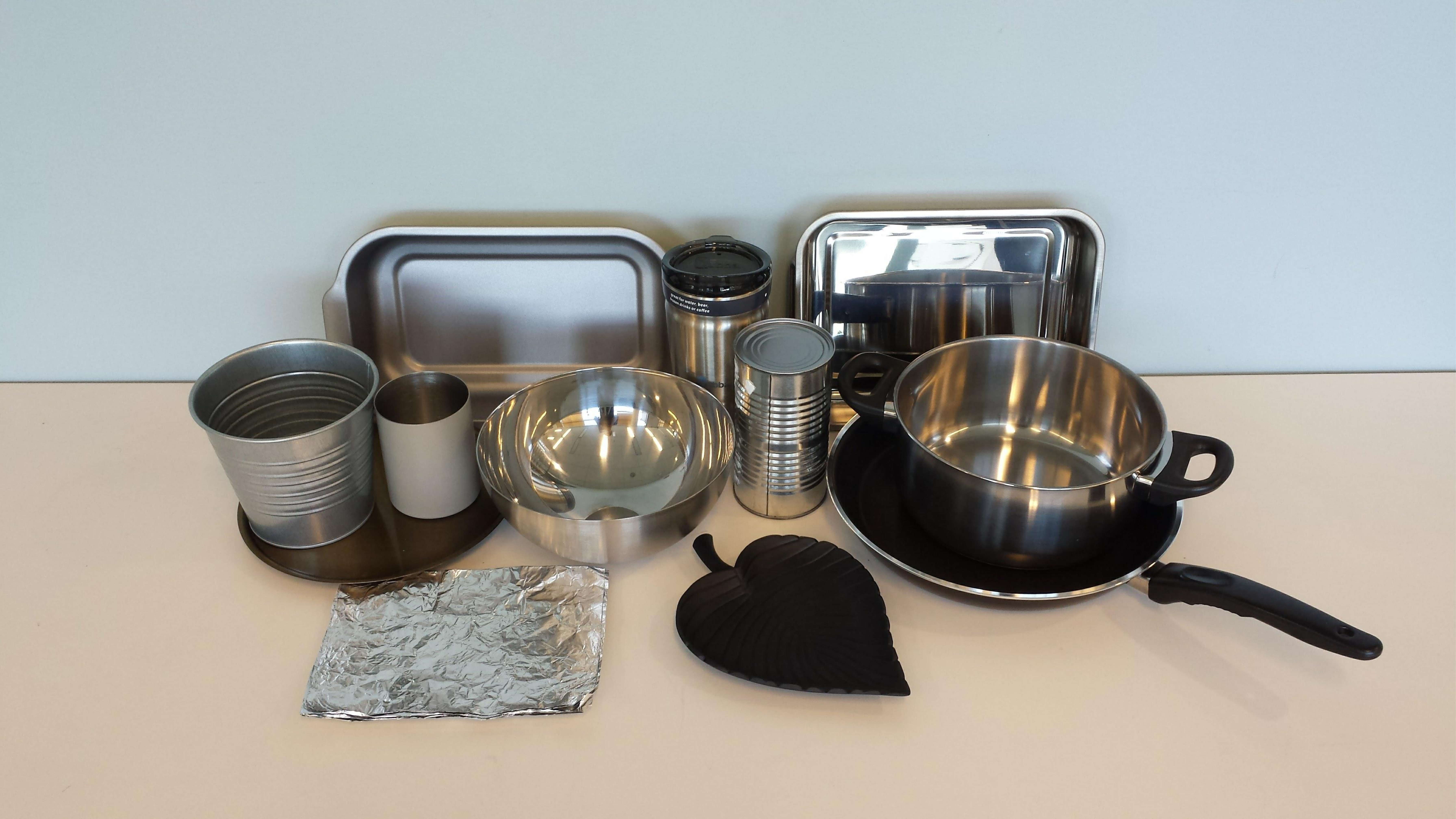}
\includegraphics[width=0.30\textwidth, trim={0cm 4cm 0cm 4cm}, clip]{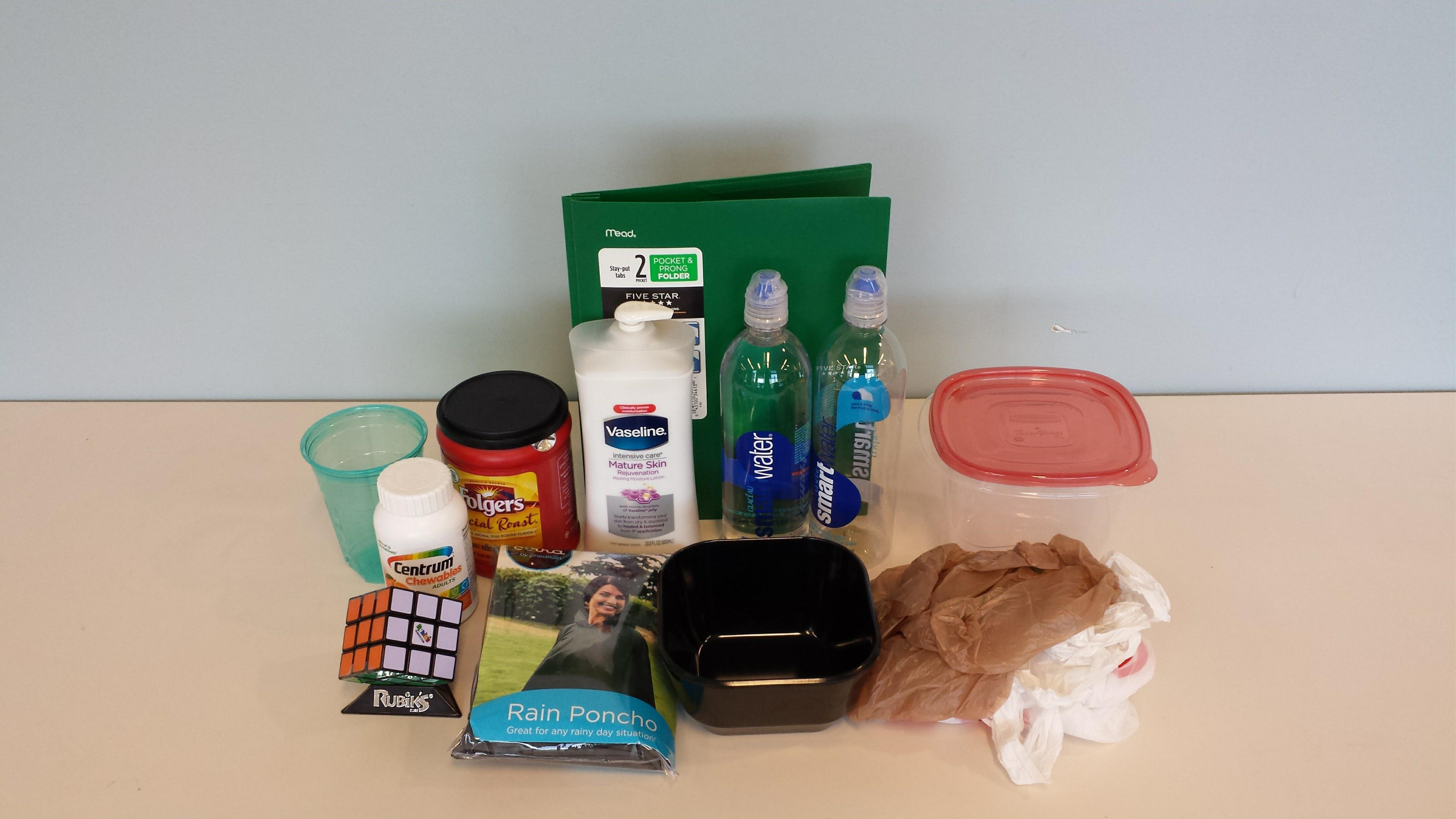}
\includegraphics[width=0.30\textwidth, trim={0cm 4cm 0cm 4cm}, clip]{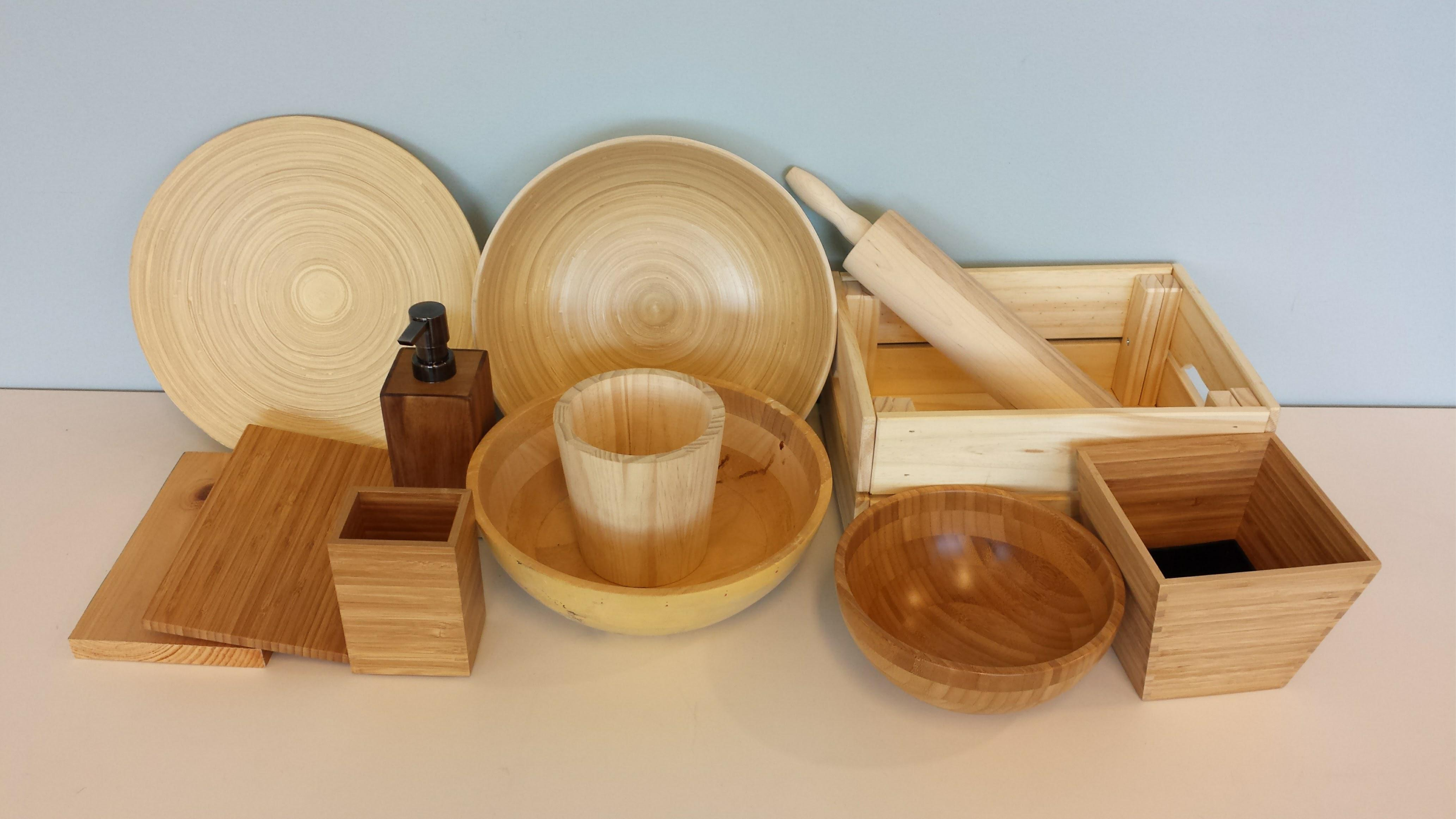}

\smallskip

\includegraphics[width=0.30\textwidth, trim={0cm 3cm 0cm 5cm}, clip]{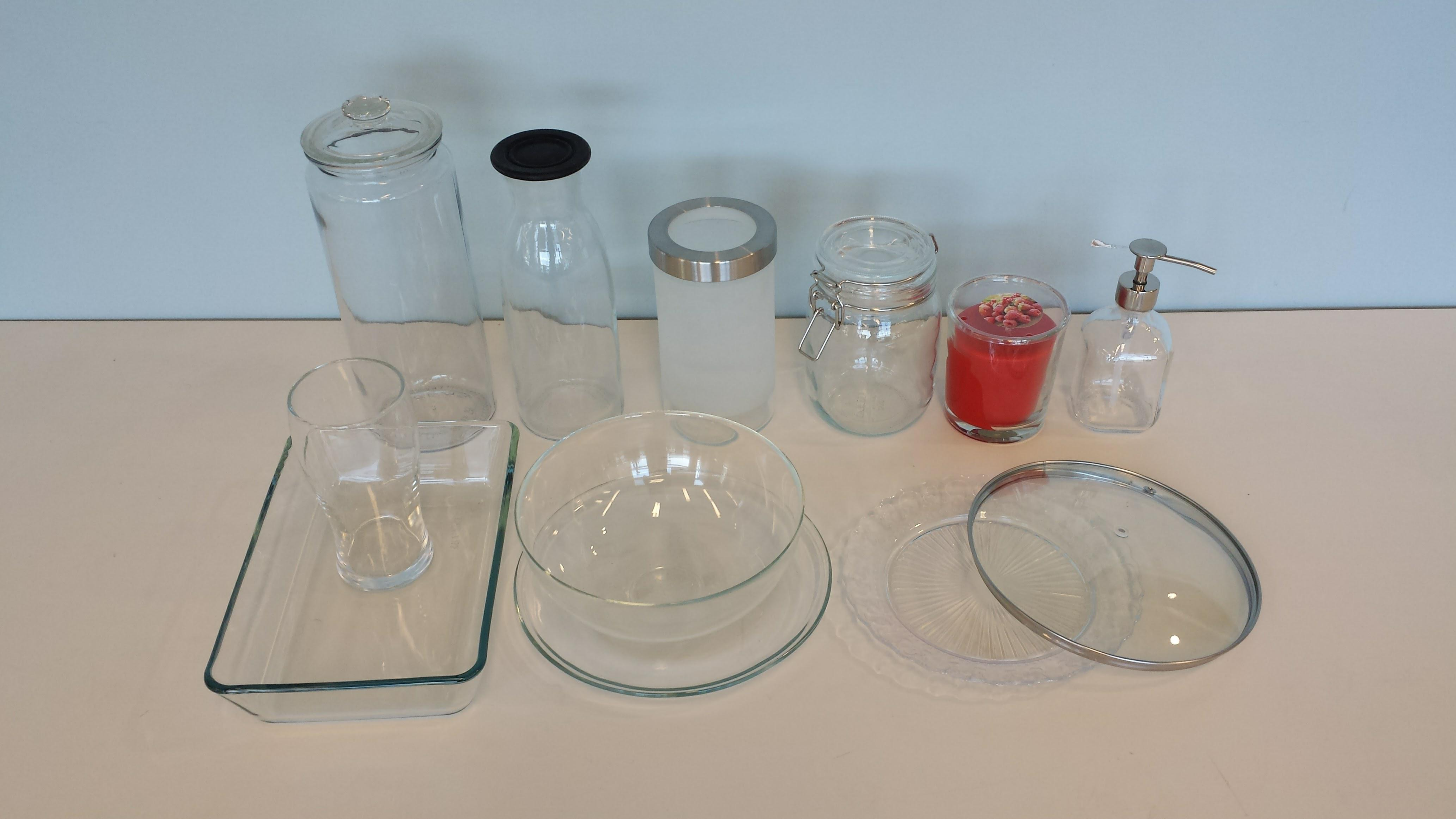}
\includegraphics[width=0.30\textwidth, trim={0cm 8cm 0cm 0cm}, clip]{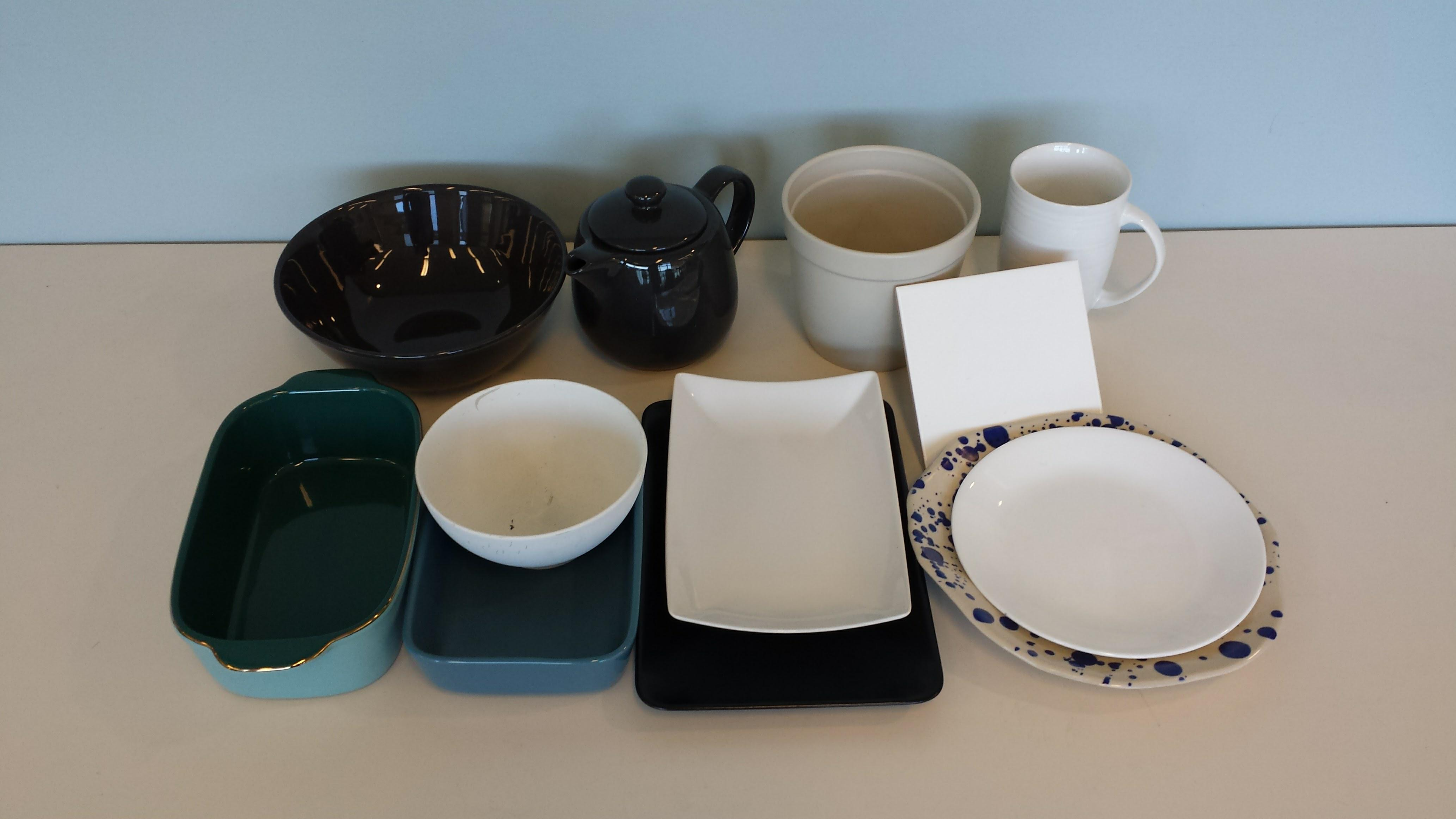}
\includegraphics[width=0.30\textwidth, trim={0cm 0cm 0cm 8cm}, clip]{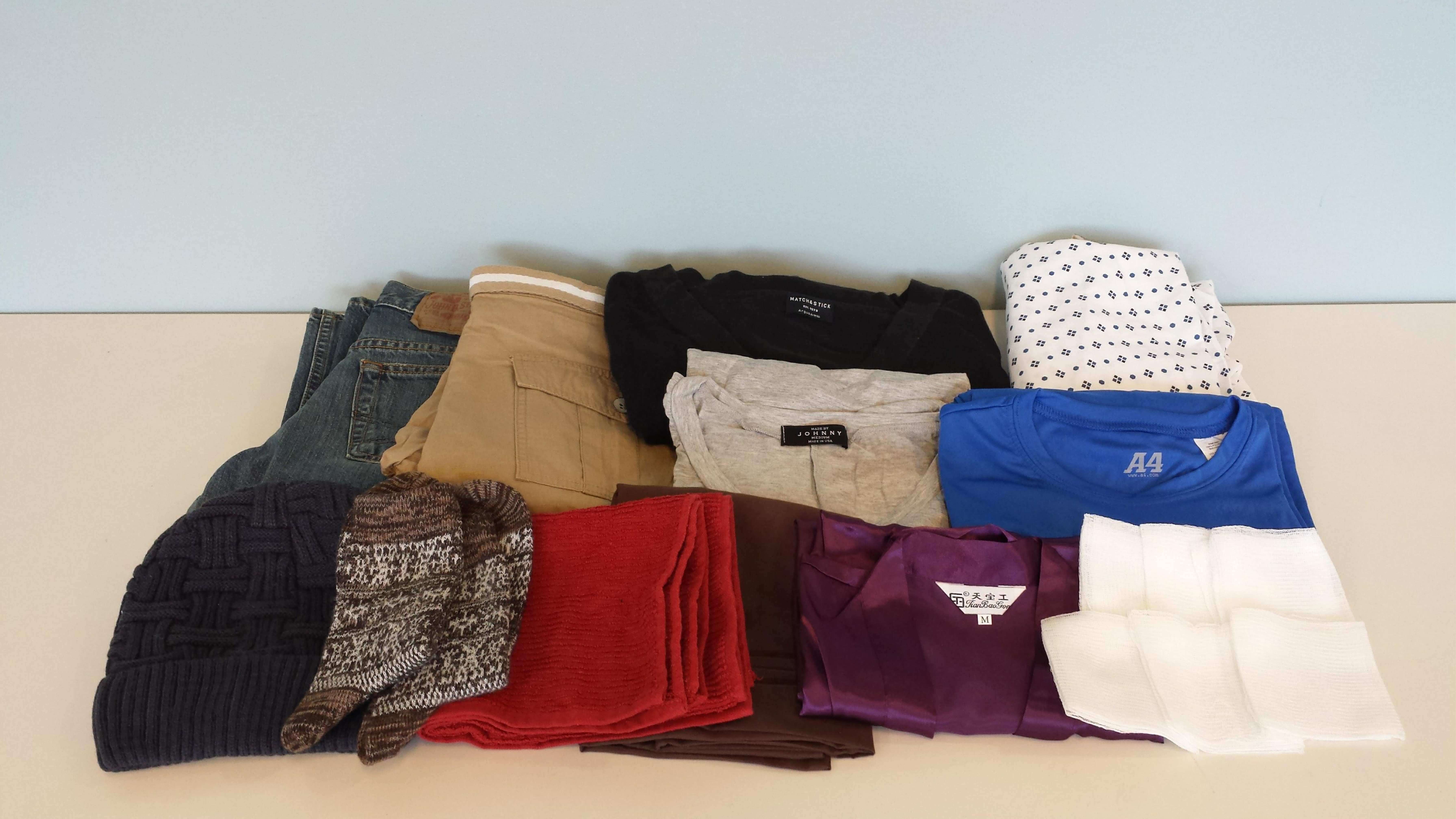}
\caption{\label{fig:imagewall}Each of the six material categories with 12 objects per category. From left to right, top to bottom: metal, plastic, wood, glass, ceramic, and fabric.}
\vspace{-1em}
\end{figure}


\section{Material Recognition of Everyday Objects (MREO) Dataset}
\label{sec:dataset}

We are releasing a dataset collected from thousands of robot interactions with household objects to support new learning approaches for material recognition. We collected our dataset on a PR2 robot that interacted with 72 objects. We distributed objects across six material categories frequently found in household environments---metal, plastic, wood, glass, ceramic, and fabric. Unlike many past works in material recognition, we do not use flat blocks of materials. We instead use objects which can be commonly found in a household, such as cups, bowls, jars, and clothing. There are 12 uniquely shaped objects per material, as shown in Figure~\ref{fig:imagewall}. 

\begin{figure}
\centering
\includegraphics[width=0.32\textwidth, trim={10cm 5cm 15cm 10cm}, clip]{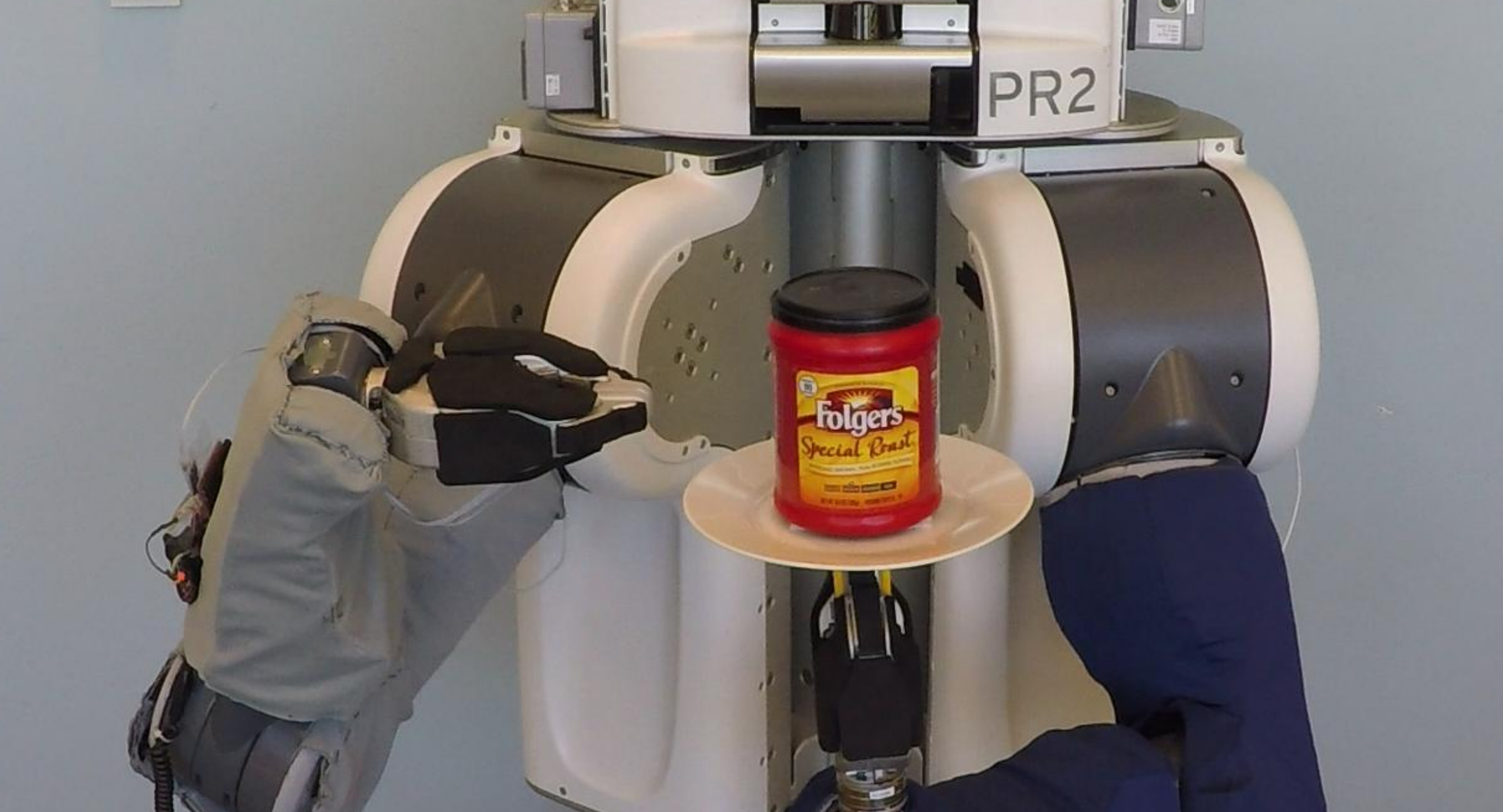}
\includegraphics[width=0.32\textwidth, trim={10cm 5cm 15cm 10cm}, clip]{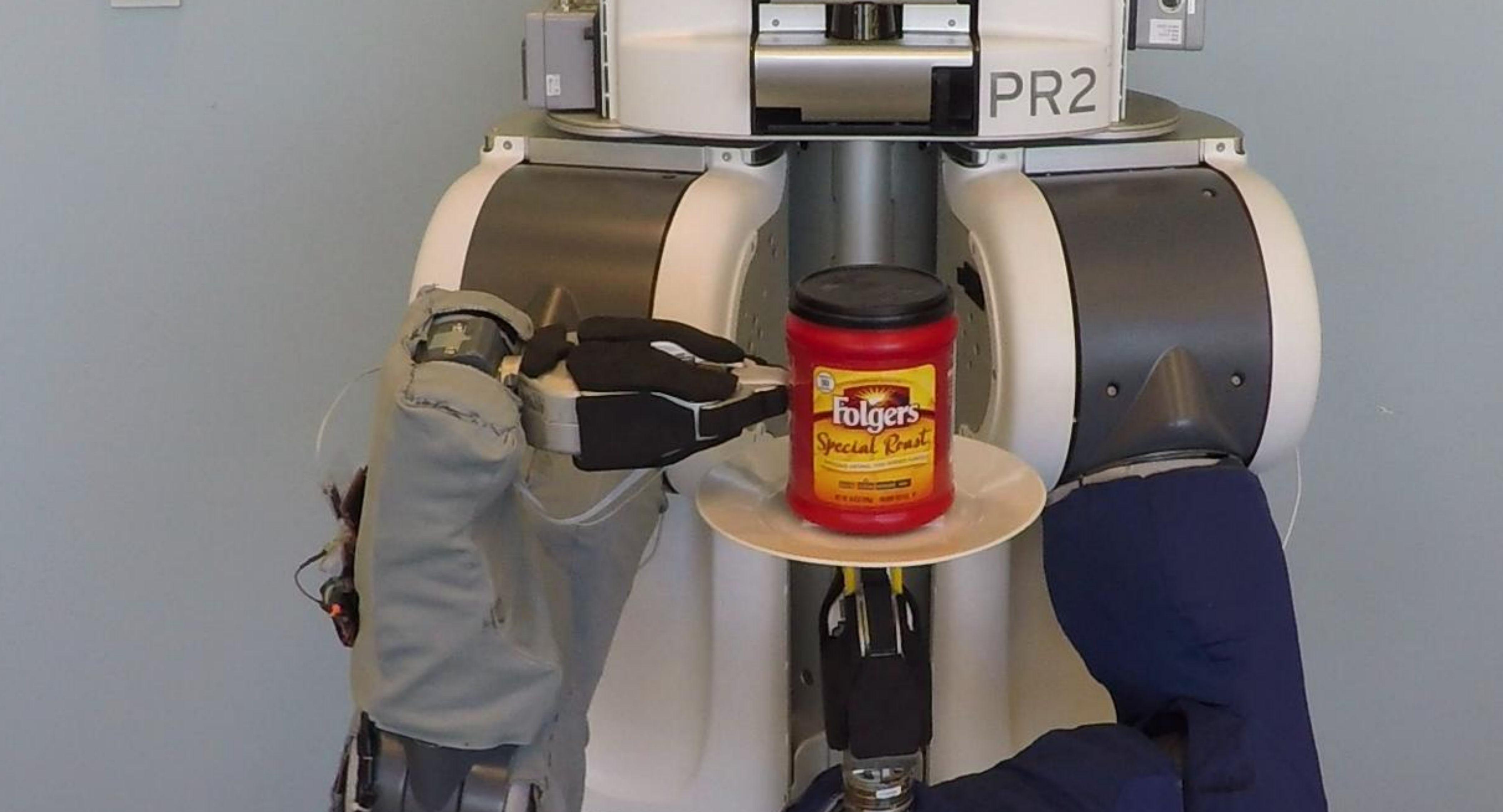}
\includegraphics[width=0.32\textwidth, trim={10cm 5cm 15cm 10cm}, clip]{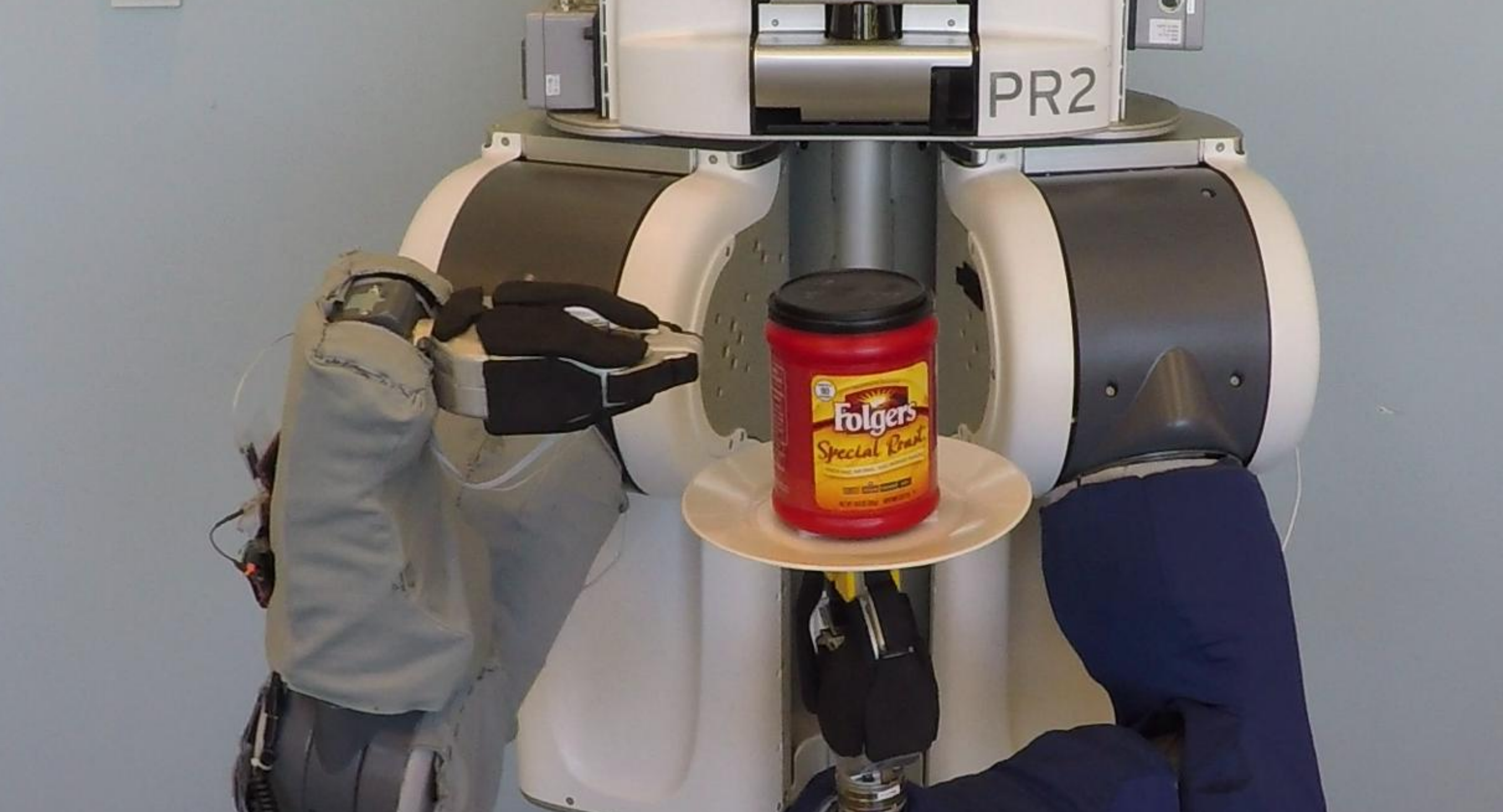}

\smallskip

\includegraphics[width=0.32\textwidth, trim={10cm 5cm 15cm 10cm}, clip]{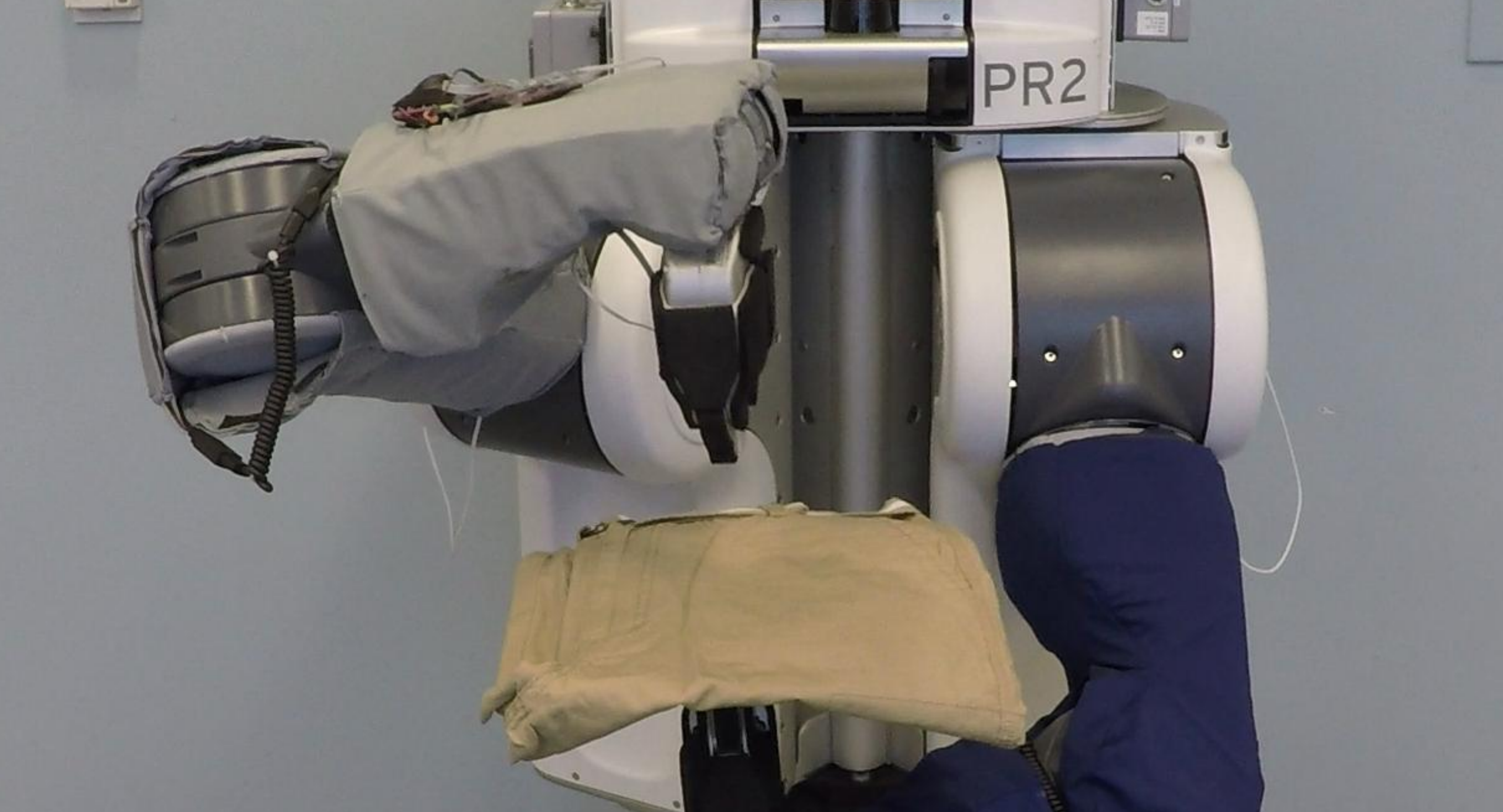}
\includegraphics[width=0.32\textwidth, trim={10cm 5cm 15cm 10cm}, clip]{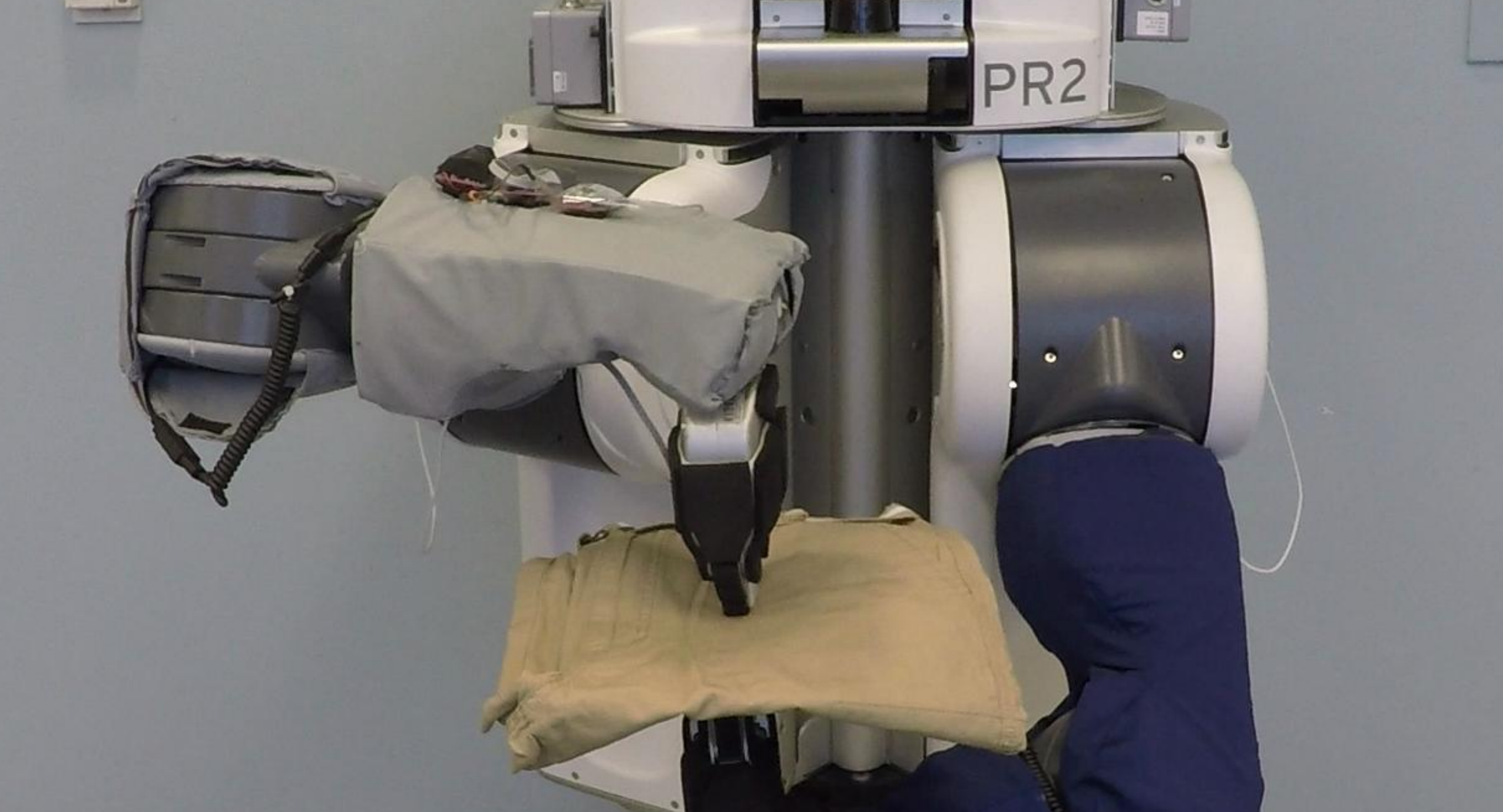}
\includegraphics[width=0.32\textwidth, trim={10cm 5cm 15cm 10cm}, clip]{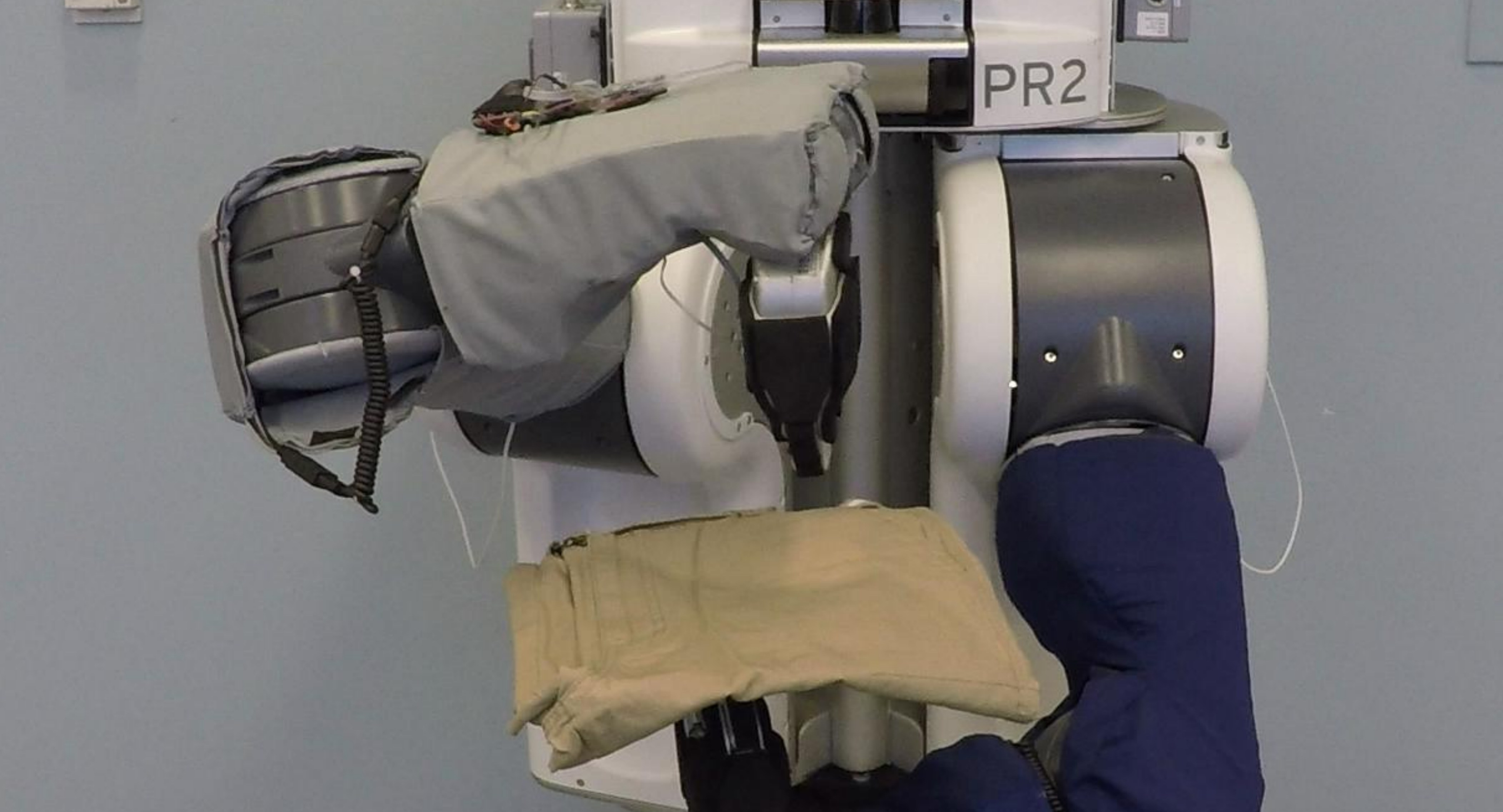}
\caption{\label{fig:interaction}The PR2 interacting with a coffee container and pair of khaki shorts through either horizontal motion (top) or vertical motion (bottom).}
\vspace{-1em}
\end{figure}

In order for the robot to interact with an object, we placed a round platter in the robot's left gripper and then rigidly attached objects to this platter. The robot then performed nonprehensile manipulation with these objects by interacting with the side or top of an object through either horizontal or vertical end effector movements. Holding a platter allowed the robot to rotate objects and interact with most of each object's exterior surface. This interaction setup is visually displayed in Figure \ref{fig:interaction}. For each interaction, the robot's end effector follows a linear trajectory using a PID controller. We also uniformly vary the robot's right end effector velocity from 5 to 10 cm/s between interactions to incorporate variability in the amount of force applied to objects at collision. In total, the robot performed 100 interactions with each object, amounting to a dataset of 7,200 interactions. When compared to prior works in haptic perception for robots, the MREO dataset spans a wide range of household objects and it is currently one of the largest datasets publicly available.

\begin{figure}
\centering
\includegraphics[width=0.50\textwidth, trim={3cm 1cm 2cm 2.5cm}, clip]{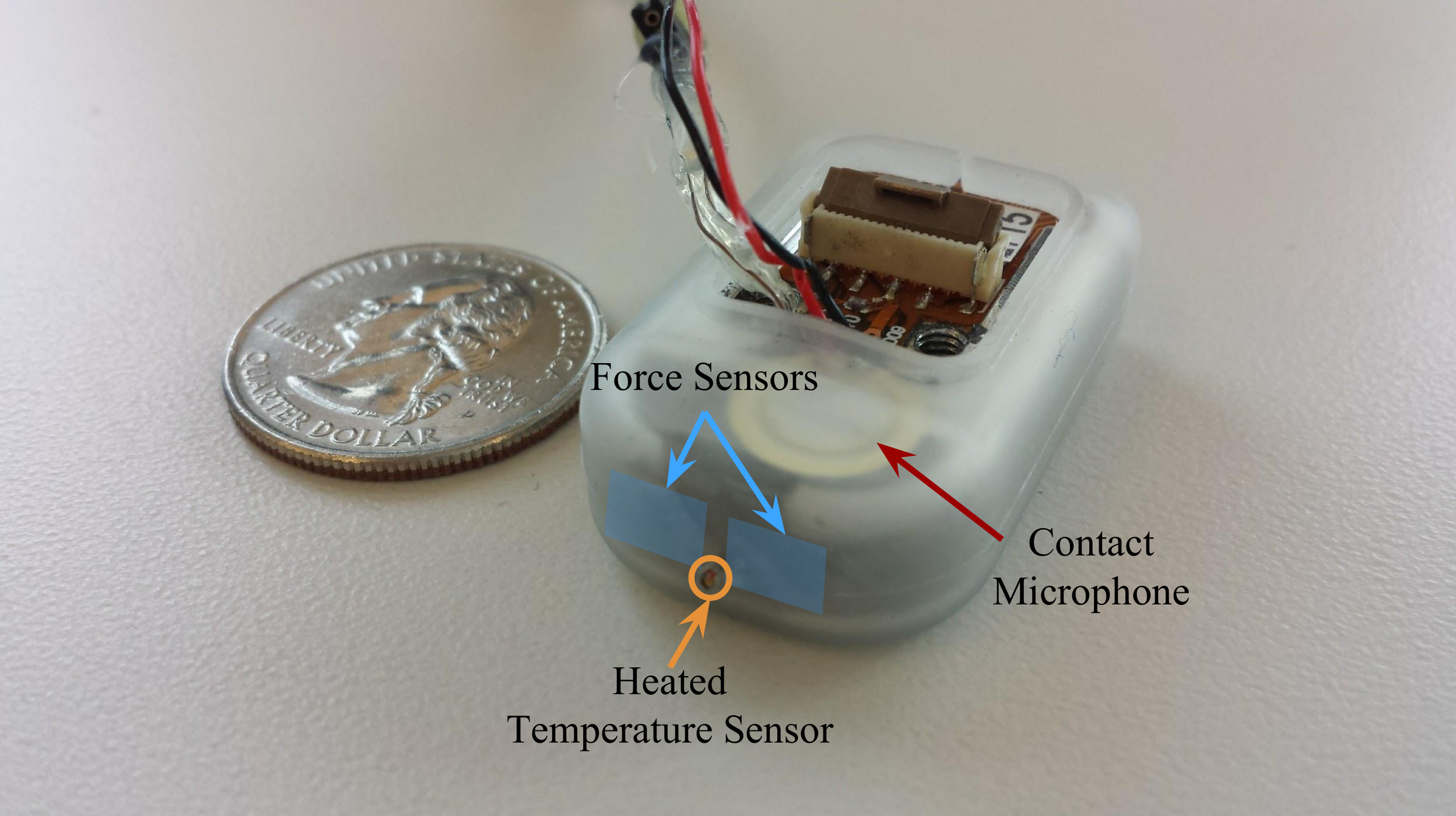}
\caption{\label{fig:fingertip}The different sensors used while interacting with objects. We use two force sensors, one actively heated temperature sensor, and one contact microphone all at the PR2's fingertip. Instructions for replicating the sensors can be found at: \url{http://healthcare-robotics.com/mr-gan}.}
\vspace{-1em}
\end{figure}

\begin{figure}
\centering
\includegraphics[width=0.32\textwidth, trim={0cm 0cm 2cm 2cm}, clip]{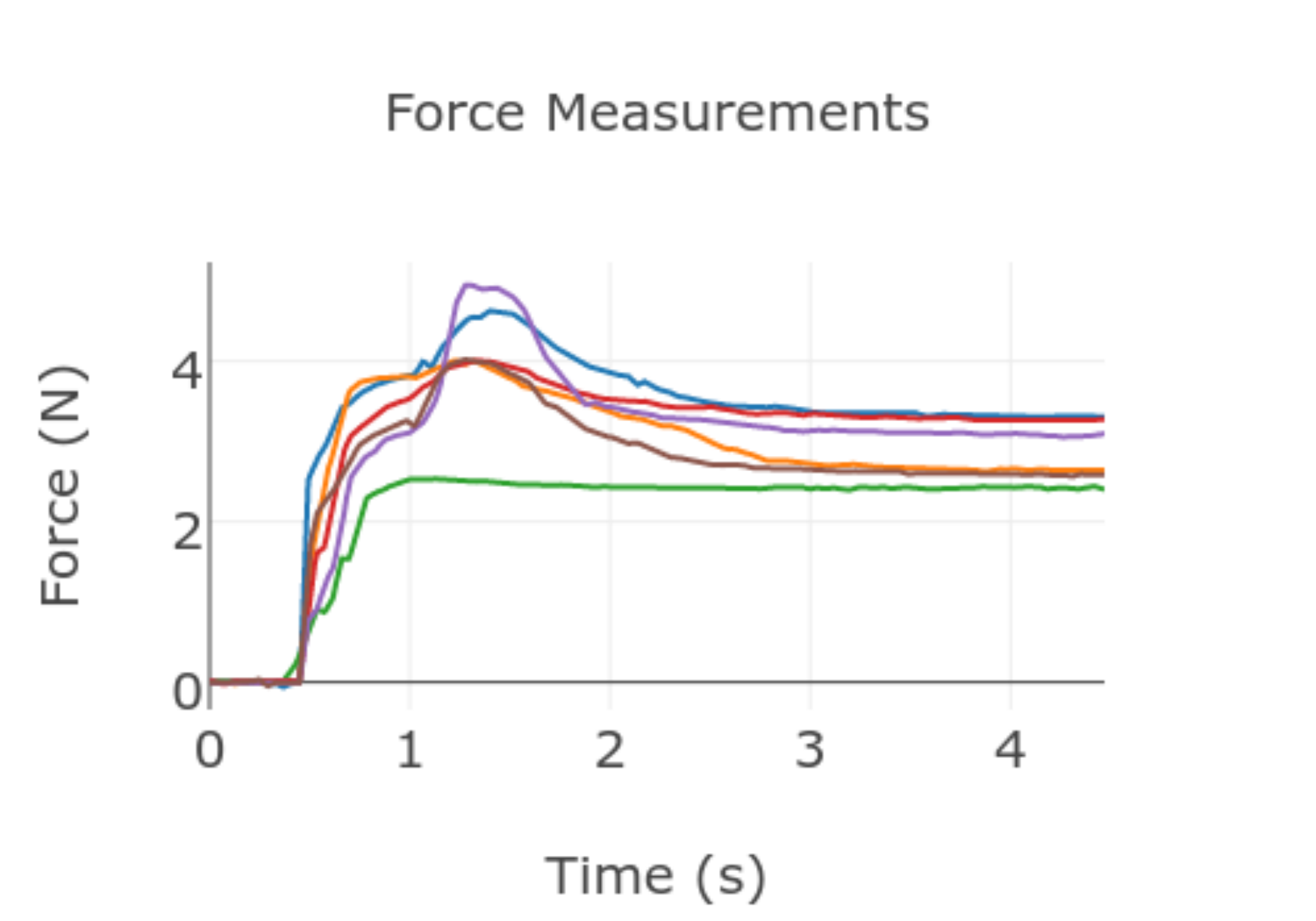}
\includegraphics[width=0.32\textwidth, trim={0cm 0cm 2cm 2cm}, clip]{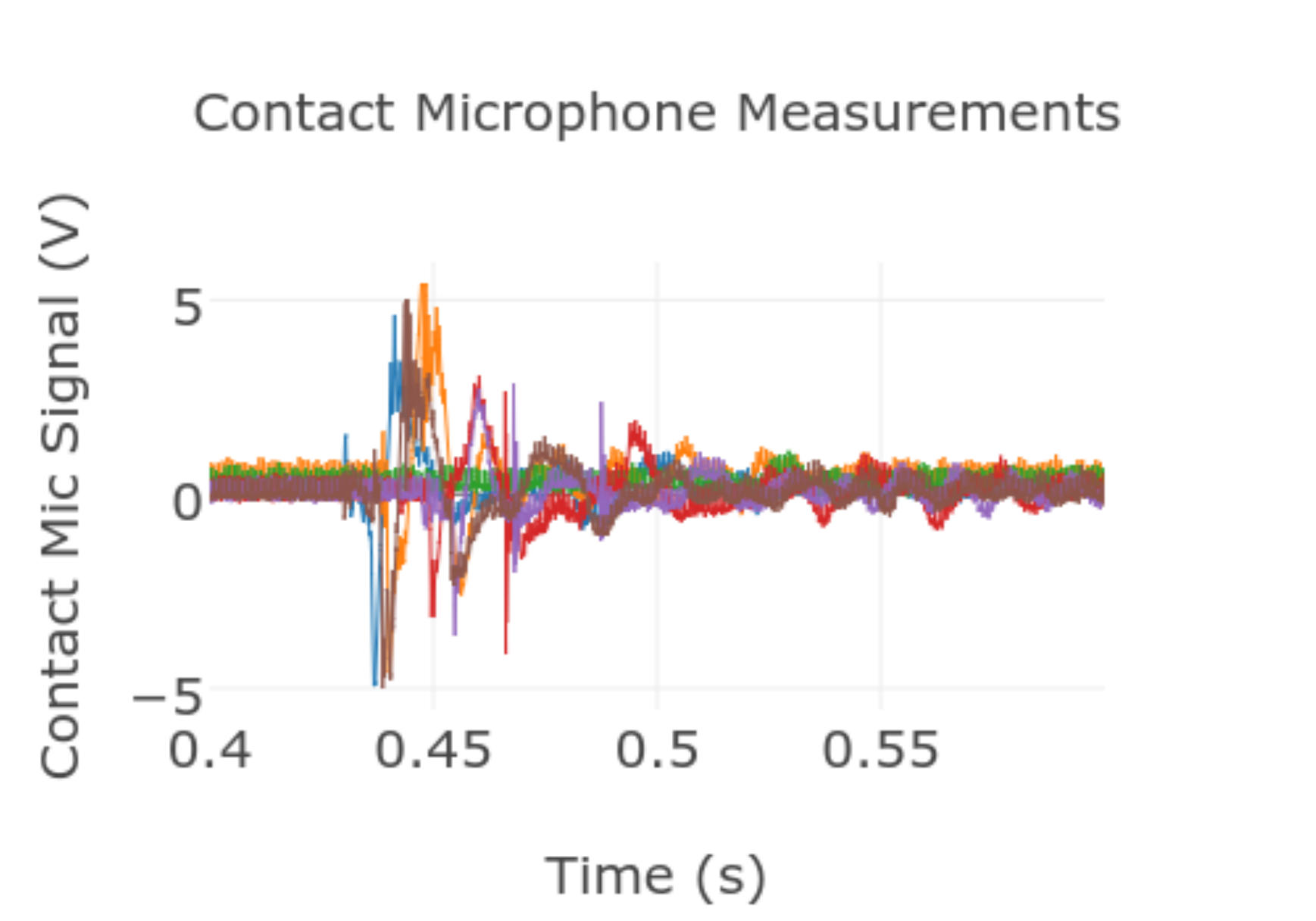}
\includegraphics[width=0.32\textwidth, trim={0cm 0cm 0cm 2cm}, clip]{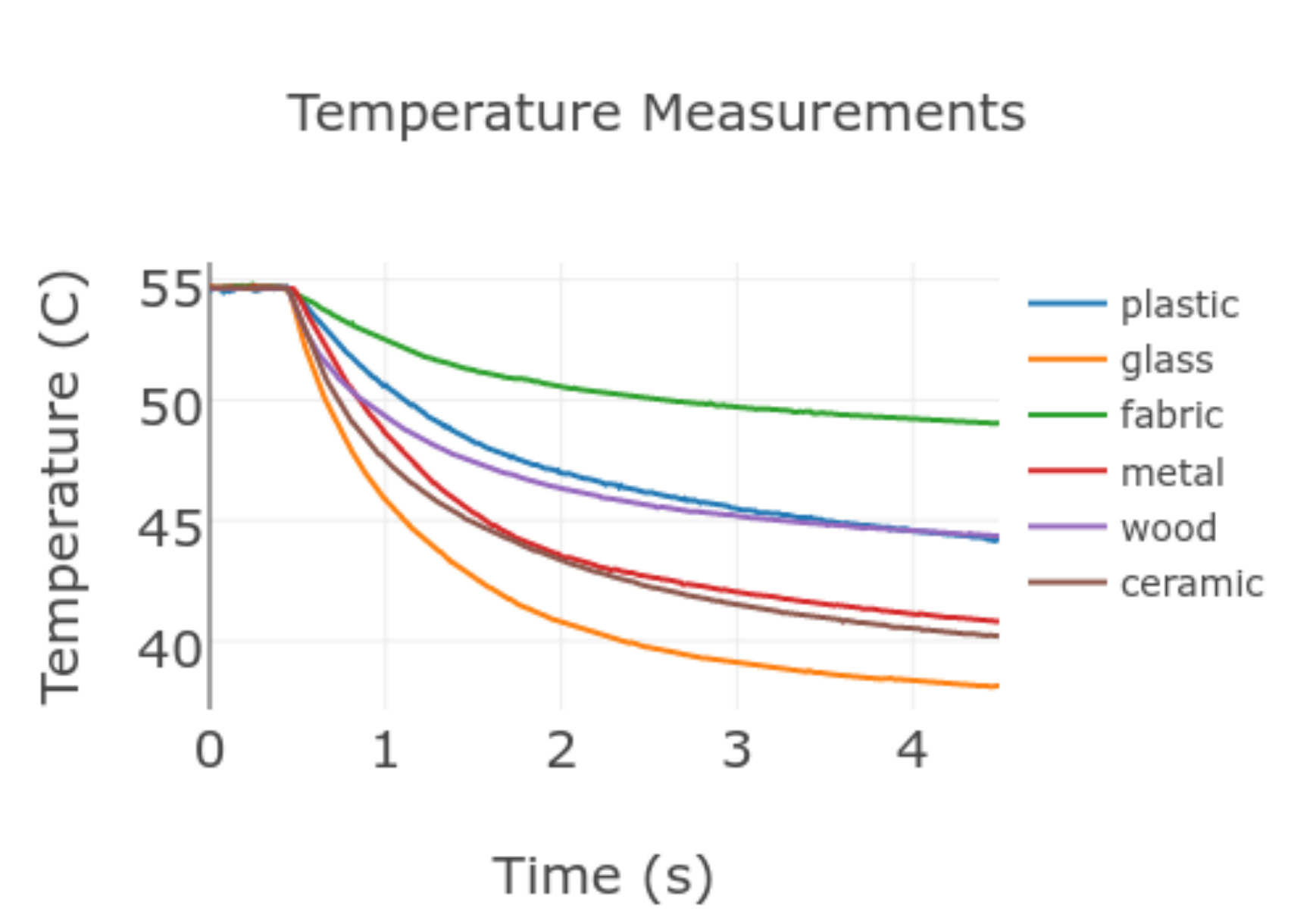}
\caption{\label{fig:signals}Example measurements from interactions with each of the six materials.}
\vspace{-1em}
\end{figure}

During each interaction, we measured time-varying signals from three different types of sensors mounted on the robot's right end effector. We used an actively heated point thermistor heated to 55 degrees Celsius prior to contact with an object. During interaction, this thermistor allows the robot to measure the rate of heat transfer into an object. Several previous works in haptics have proposed contact microphones for measuring vibrations during contact~\citep{chen2016learning, murray2008stane}. Given this, we attached a contact microphone to the PR2's fingertip to measure the vibrations that occurred at the moment of impact with an object. We also measured applied forces via two force sensors that are built into the robot's fingertip. Figure~\ref{fig:fingertip} shows all of these sensors. We chose these three modalities---force, temperature, and vibration---since all three are found in the human somatosensory system. We use force and temperature sensing feedback to halt the robot's movement when contact is made with an object. Contact was detected when either one of the force sensors exceeded a measured reading of 1 N, or the temperature sensor differed by more than 1 degree from its initial set point. During interaction, we sampled the temperature and force sensors at a rate of 100 Hz and 25 Hz, respectively. We sampled the contact microphone at $\sim$35 kHz as higher sampling rates are advantageous for capturing vibrations upon impact between the end effector and object. Figure~\ref{fig:signals} displays examples of force, contact microphone, and temperature signals collected during interactions with an object.

\section{Semi-Supervised Learning with GANs}
\label{sec:semigans}

A generative adversarial network (GAN) consists of a minimax adversarial game between a generator, $G$, and a discriminator, $D$~\citep{NIPS2014_5423}. The generator aims to output realistic data samples by mapping random noise $\mathbf{z}$ from a noise distribution $p(\mathbf{z})$ to a point in the data distribution. The discriminator is then optimized to differentiate between a real data sample, $\mathbf{x}$, and a sample from the generator, $\mathbf{\tilde{x}}=G(\mathbf{z})$. Recently, GANs have found success in a range of applications such as image generation, super-resolution imaging, video frame prediction, and image classification~\citep{radford2016unsupervised, ledig2016photo, mathieu2015deep, salimans2016improved}. Several works have also extended GANs into semi-supervised learning for image classification~\citep{Makhzani2016, springenberg2015unsupervised}.

A typical multiclass classifier will label a new data point as one of $K$ discrete classes. For semi-supervised learning, we can add samples from the generator $G$ into our dataset with a new label of $K+1$, corresponding to a ``fake" class. As described by~\citet{salimans2016improved}, we can learn from unlabeled data by defining a loss function for the discriminator as,
\begin{align}
	\label{eq:1}
	L &= L_{\text{supervised}} + L_{\text{unsupervised}} \\
	L_{\text{supervised}} &= -\mathop{\mathbb{E}}\nolimits_{\mathbf{x},y\sim p_{\text{data}}(\mathbf{x}, y)}\log p_{\text{model}}(y|\mathbf{x}, y<K+1) \notag \\
    L_{\text{unsupervised}} &= -\mathop{\mathbb{E}}\nolimits_{\mathbf{x}_u\sim p_{\text{data}}(\mathbf{x}_u)}\log(1\! -\! p_{\text{model}}(y\! =\! K\! +\! 1|\mathbf{x}_u)) - \mathop{\mathbb{E}}\nolimits_{\mathbf{\tilde{x}}\sim G(\mathbf{z})}\log p_{\text{model}}(y\! =\! K\! +\! 1|\mathbf{\tilde{x}}). \notag
\end{align}

Here, $p_{\text{model}}(y=K+1|\mathbf{x})$ represents the probability that $\mathbf{x}$ is fake and corresponds to $1-D(\mathbf{x})$ from the original GAN architecture~\citep{NIPS2014_5423}. $\mathbf{x}_u$ denotes an unlabeled data sample. We can train the classifier with gradient descent by minimizing both $L_{\text{supervised}}$ and $L_{\text{unsupervised}}$. Intuitively, $L_{\text{supervised}}$ emphasizes correctly classifying a material given labeled sensor measurements, $x$. The first term in $L_{\text{unsupervised}}$, focuses on classifying unlabeled sensor measurements as one of the $K$ classes, whereas the second term aims to classify images from the generator as fake, i.e. $K+1$.

At each training iteration, we stochastically update the discriminator by descending the gradient of equation~\ref{eq:1}:
\[
\nabla_{\theta_d} \frac{1}{m}\sum\limits^m_{i=1} -\log \sigma(\mathbf{x}^{(i)})_{y^{(i)}} - \log D(\mathbf{x}^{(i)}_u) - \log (1 - D(G(\mathbf{z}^{(i)}))),
\]
where $\sigma(\mathbf{x})_j = p_{\text{model}}(y=j|\mathbf{x})$ is the softmax function applied to the output of the discriminator for all $m$ samples in a minibatch. \citet{salimans2016improved} proposed a technique, feature matching, to improve classification performance by defining a new cost function for the generator that discourages it from overtraining on the discriminator. We update the generator with feature matching by descending its gradient given by the activations, $\mathbf{f}(\mathbf{x})$, on an intermediate layer of the discriminator:
\[
\nabla_{\theta_g} \left\lVert\frac{1}{m}\sum\limits^m_{i=1} \mathbf{f}(\mathbf{x}^{(i)}_u) - \mathbf{f}(G(\mathbf{z}^{(i)})) \right\rVert^2_2
\]

Our generator has three fully connected neural network layers with a softplus activation function applied to the output of the first two layers and batch normalization between the first and second layer. The generator takes as input a 100-dimensional noise vector sampled from a Gaussian distribution and outputs a vector corresponding to a generated time series of haptic signals.

Our discriminator has six fully connected layers with ReLU activations and Gaussian noise added to the output of each hidden layer for regularization. The vector output of the discriminator can be mapped to a single class estimate by applying an argmax function. We use the Adam optimizer \citep{kingma2014adam} while training both the generator and discriminator with a learning rate of 0.0006 and $\beta_1=$ 0.5. 

\section{Evaluation}
\label{sec:evaluation}

We conducted several semi-supervised experiments to evaluate the performance of our generative adversarial approach given unlabeled training samples and new objects. We measured sensor data as the robot interacted with each object, as discussed in Section~\ref{sec:dataset}. Prior to learning, we mapped these unprocessed data to vectors of fixed length via linear interpolation over each sensor signal with uniform resampling. We interpolated force and temperature signals to match a sampling frequency of 100 Hz. We interpolated the contact microphone signals to a high quality audio frequency of 48 kHz and then computed the input features as a Mel-scaled spectrogram. We used 128 bands for the Mel filter bank which separates the signal into multiple single-frequency components. Similar to \citet{deng2013recent}, we found that Mel-scaled spectrogram features resulted in stronger classification performance than did raw vibration signals. One second of vibration data resulted in a vector of size 48,000, which we converted to a Mel-scale spectrogram of length 12,032 with 128 frequency bands. 

\subsection{Semi-Supervised Learning}
\label{sec:semisupervised}

For semi-supervised learning, we evaluate our GAN's classification accuracy given varying amounts of labeled and unlabeled training data. Our dataset consists of 7,200 recorded time series, with 1,200 time series per material. We computed results via stratified 6-fold cross-validation. Each fold had 1,200 samples consisting of 200 randomly selected time series from each of the six material classes.

For temperature and force features, we considered only measurements between 0.1 seconds before the robot's gripper contacts an object and 4 seconds after contact. Similarly, we only used measurements from the contact microphone between 0.1 seconds before contact and 0.1 seconds after contact. We trained each model over 100 epochs.

Table~\ref{table:semisupervised_results} summarizes our results. 1\% of labeled training data amounts to providing 10 labeled time series from each of the six classes, with the other 990 provided without labels. We place the remaining 200 time series from each class into the testing fold. With all three modalities and all training data labeled, our classifier can recognize materials with 96.2\% accuracy. Our classifier achieves over 95\% accuracy when half of the training data were unlabeled and almost 90\% accuracy even when 92\% of the data were unlabeled. The classification performance tended to drop more rapidly when there were fewer than 80 labeled samples per class and this remains an area for future research. Surprisingly, a contact microphone actively hurt material recognition when less than 4\% of the training data were labeled. We suspect this is due to variability in the high frequency vibration signals which makes these signals hard to disambiguate without enough labeled samples. 

We also compared these results to that of supervised learning by training on only labeled data and disregarding any unlabeled data. We compared against both an SVM with a radial basis function (RBF) kernel and a fully-connected neural network, which used the discriminator's architecture from our GAN. Results are shown in Table~\ref{table:svmnn_results}. When less than half of the data are labeled, semi-supervised learning can improve performance by up to 10\% over a supervised SVM or neural network.

\begin{table}
\centering
\caption{\label{table:semisupervised_results}Semi-supervised learning results. Results are presented as accuracies in percent and are averaged over six test sets via 6-fold cross validation. Each column represents the percentage of training data that was labeled during training, with the remaining data provided without labels.}
\begin{tabular}{cccccccc} \toprule 
    & \multicolumn{7}{c}{Percentage of training data labeled} \\ \cmidrule{2-8}
    Modalities & 1\% & 2\% & 4\% & 8\% & 16\% & 50\% & 100\% \\ \midrule\midrule
    Force & 62.1 & 70.4 & 72.2 & 77.7 & 79.8 & 85.8 & 87.9 \\
    Temperature & 53.8 & 59.0 & 64.1 & 68.1 & 69.0 & 80.0 & 82.1 \\
    Contact mic & 42.9 & 53.9 & 62.6 & 67.5 & 73.4 & 79.8 & 83.1 \\
    Force, Temperature & \textbf{74.3} & \textbf{81.4} & \textbf{85.6} & 88.5 & 90.2 & 94.2 & 95.3 \\
    Force, Contact mic & 58.2 & 67.5 & 73.8 & 80.2 & 84.7 & 89.7 & 91.8 \\
    Temperature, Contact mic & 52.4 & 68.3 & 79.2 & 84.9 & 87.4 & 91.2 & 92.2 \\
    Force, Temperature, Contact mic & 62.8 & 75.4 & \textbf{85.6} & \textbf{89.4} & \textbf{92.0} & \textbf{95.4} & \textbf{96.2} \\ 
	\bottomrule
\end{tabular}
\end{table}

\begin{table}
\centering
\caption{\label{table:svmnn_results}Classification results with supervised learning using both an SVM and the discriminator network from the proposed GAN architecture. This compares directly with Table~\ref{table:semisupervised_results} by training the SVM and fully-connected network on only the labeled training data (disregarding unlabeled data).}
\begin{tabular}{cccccccc} \toprule 
     & \multicolumn{7}{c}{Percentage of dataset used for training} \\ \cmidrule{2-8}
    Supervised SVM with RBF Kernel & 1\% & 2\% & 4\% & 8\% & 16\% & 50\% & 100\% \\ \midrule\midrule
    Force, Temperature & 60.4 & 64.6 & 69.6 & 74.1 & 78.2 & 86.0 & 88.8 \\
    Force, Temperature, Contact mic & 58.9 & 64.6 & 70.4 & 75.8 & 80.5 & 88.9 & 91.8 \\ \midrule \\
    Supervised Discriminator & 1\% & 2\% & 4\% & 8\% & 16\% & 50\% & 100\% \\ \midrule\midrule
    Force, Temperature & 62.9 & 71.3 & 78.6 & 83.3 & 87.0 & 91.9 & 93.6 \\
    Force, Temperature, Contact mic & 59.1 & 69.8 & 78.6 & 85.8 & 89.7 & 92.6 & 94.8 \\
    \bottomrule
\vspace{-2em}
\end{tabular}
\end{table}

\subsection{Generalizing to New Objects}

When robots begin entering real world environments, they will likely be faced with new objects that they have not interacted with before. As such, we also evaluated how well our generative model performs when classifying objects that are not found in the training set. We assessed generalization over a wide variety of objects via leave-one-object-out cross-validation. For this, we trained our model on 71 objects (7,100 interactions) and computed classification accuracy on the remaining 100 interactions from the left out object. We repeated this process for every object and report the resulting accuracy as the average over all 72 training and test sets. We followed a similar training structure as specified above, using the same duration of contact data and training for 100 epochs. 

Table~\ref{table:generalization_results} shows our generalization results using leave-one-object-out cross-validation with semi-supervised training. Our approach achieved an accuracy of 75.1\% with all three modalities and all training sets fully labeled. We note that generalization is challenging for our dataset since most objects are fairly distinct from one another, as seen in Figure~\ref{fig:imagewall}. This variation is further exaggerated as many objects have paint or coatings, such as the metal frying pan which is the only object with a Teflon coating. We again compare results to supervised learning with an SVM and a fully-connected discriminator, shown in Table~\ref{table:generalization_svmnn}. For generalization, we observe that semi-supervised learning with a GAN provides little benefit over supervised learning. This remains an open area for further research.

Overall, we found that objects with odd or uneven surface contours posed the largest problem for generalization as these contours could lead to unbalanced force distributions among the two force sensors. Similarly, the temperature sensor could slip along a highly curved surface and give inconsistent readings between interactions. For instance, our models had difficulty generalizing classification to the glass and plastic bowls due to their highly curved surfaces. However, our models generalized well to objects such as clothing articles, large cylindrical glass jars, and objects with flat surfaces.


\begin{table}
\centering
\caption{\label{table:generalization_results}Generalization results via leave-one-object-out cross-validation. Each result is presented as the average accuracy over 72 GAN models, where each model is trained on 71 objects and tested on the remaining object.}
\begin{tabular}{cccccc} \toprule
	Leave-One-Object-Out & \multicolumn{5}{c}{Percentage of training data labeled} \\ \cmidrule{2-6}
    Modalities & 1\% & 4\% & 16\% & 50\% & 100\% \\ \midrule\midrule
    Force, Temperature & 65.4 & 68.1 & 69.7 & 68.4 & 70.3 \\
    Force, Temperature, Contact mic & 56.4 & 67.0 & 70.6 & 73.6 & 75.1 \\
    \bottomrule
\vspace{-1em}
\end{tabular}
\end{table}

\begin{table}
\centering
\caption{\label{table:generalization_svmnn}Generalization results for supervised learning with an SVM and fully-connected discriminator model. Results are computed via leave-one-object-out cross-validation.}
\begin{tabular}{cccccc} \toprule
	Leave-One-Object-Out & \multicolumn{5}{c}{Percentage of daataset used for training} \\ \cmidrule{2-6}
    Supervised SVM with RBF Kernel & 1\% & 4\% & 16\% & 50\% & 100\% \\ \midrule\midrule
    Force, Temperature & 50.9 & 61.5 & 67.5 & 70.7 & 72.6 \\
    Force, Temperature, Contact mic & 50.5 & 59.6 & 68.0 & 72.7 & 75.1 \\ \midrule \\
    Supervised Discriminator & 1\% & 4\% & 16\% & 50\% & 100\% \\ \midrule\midrule
    Force, Temperature & 58.3 & 66.1 & 69.3 & 72.1 & 74.9 \\
    Force, Temperature, Contact mic & 54.8 & 68.5 & 74.7 & 74.6 & 76.6 \\
    \bottomrule
\vspace{-2em}
\end{tabular}
\end{table}

\subsection{Varying Duration of Interaction}
\label{sec:duration}

We also investigated how the duration of contact with an object impacts material recognition. Several past studies, such as work by \citet{kerr2013material} and \citet{takamuku2008robust}, have shown that strong material recognition results can be achieved when the robot remains in contact with an object for over 20 seconds. However, these long duration contact requirements reduce the rate at which a robot can interact with and sense its surroundings. To address this issue, we compared how GAN models performed using different sensing modalities and decreasingly short durations of contact.

The overall testing procedure, data processing, and network training remained similar to that of semi-supervised learning in Section~\ref{sec:semisupervised}. However, we provided labels for all training data and instead varied the duration of data that we considered once the robot made contact with an object. Table~\ref{table:lengths_results} summarizes our results when we trained our GAN architecture on various lengths of interaction.

With 0.5 seconds of temperature and force data, our method achieved a classification accuracy of 92.4\% which is just 2.9\% lower than when the robot interacted with an object for an entire 4 seconds. As seen in Figure~\ref{fig:signals}, temperature and force signals began to stabilize after 2 seconds of contact, thus leading to no significant performance gains with longer durations of contact. 

When contacting an object for 0.2 seconds, our approach was 8\% more accurate with a contact microphone than either force or temperature sensing alone. Therefore, high frequency vibrations may be useful for rapid classification upon contact with an object. For example, these vibrations may give a robot insight into the material of an object that it briefly bumps into when manipulation in cluttered spaces. However, contact microphones also have limitations. Namely, these high frequency vibrations lead to large feature vectors which are more difficult to process and evaluate in real time.

\begin{table}
\centering
\caption{\label{table:lengths_results}Classification accuracy as the length of interaction with objects varies. Results are presented in percent and are computed with 100\% of the training data labeled. While one second of contact microphone data provided higher accuracy for supervised learning, we found that 0.2 seconds of data improved results during semi-supervised learning.}
\begin{tabular}{cccccccc} \toprule
    & \multicolumn{7}{c}{Length of interaction in seconds} \\ \cmidrule{2-8}
    Modalities & 4s & 3s & 2s & 1s & 0.5s & 0.2s & 0.1s \\ \midrule\midrule
    Force & 87.9 & 87.6 & 87.6 & 86.9 & 81.8 & 75.1 & 70.9 \\
    Temperature & 82.1 & 80.3 & 77.5 & 73.9 & 70.4 & 64.4 & 58.9 \\
    Force, Temperature & \textbf{95.3} & 94.8 & 95.0 & 94.4 & 92.4 & 88.6 & 84.4 \\ \midrule \\
    & 1s & 0.7s & 0.5s & 0.3s & 0.2s & 0.1s & 0.05s \\ \midrule\midrule
    Contact mic & \textbf{84.6} & 84.0 & 83.8 & 82.4 & 83.1 & 77.0 & 63.3 \\
    \bottomrule
\vspace{-1em}
\end{tabular}
\end{table}

\begin{table}
\centering
\caption{\label{table:unlabeled_data_results}Material recognition accuracy as the amount of unlabeled training data increases. We provided only 40 labeled samples per class during training. The far right column with 960 unlabeled training samples corresponds to the 4\% column in Table~\ref{table:semisupervised_results}.}
\begin{tabular}{cccccccc} \toprule
	40 labeled samples per class & \multicolumn{7}{c}{Number of unlabeled training samples per class} \\ \cmidrule{2-8}
    Modalities & 0 & 40 & 80 & 160 & 320 & 640 & 960 \\ \midrule\midrule
    Force, Temperature & 74.8 & 77.5 & 78.5 & 81.1 & 83.0 & 85.8 & 85.6 \\
    Force, Temperature, Contact mic & 71.9 & 72.4 & 76.4 & 79.3 & 81.7 & 83.2 & 85.6 \\ 
    \bottomrule
\vspace{-2em}
\end{tabular}
\end{table}

\subsection{Impact of Unlabeled Data}
\label{sec:unlabeled_impact}
 
Once deployed in the real world, it may be valuable for a robot to continue training on unlabeled data that it observes when interacting with objects. Yet, it is unclear if providing more unlabeled data during training will lead to any improvements in performance. To address this, we trained models on increasing quantities of unlabeled data, while holding the amount of labeled training data constant.

We followed a similar training setup as in Section~\ref{sec:semisupervised} with 6-fold cross-validation and 100 epochs of training. However, for each training fold we provided 40 randomly selected labeled training examples for each class (4\% of the entire training set) and then varied the amount of unlabeled data provided during training. Table~\ref{table:unlabeled_data_results} presents our results.

With just force and temperature features, our GAN achieved 74\% accuracy when trained with no unlabeled data and only 40 labeled samples per material class. With 40 unlabeled samples per class we see a classification improvement of just 2.7\%. However, with 960 unlabeled training examples per class---24 times more unlabeled data than labeled---material recognition accuracy increases by over 10\%. Given these results, it seems reasonable that a robot could improve its performance by continuing to train on unlabeled data that it observes when operating in everyday environments.
 

\section{Conclusion}

We describe how a robot can use generative adversarial networks to learn about material recognition when most of the provided training data are unlabeled. Our robot uses force, temperature, and vibration features in order to differentiate objects given six material categories. Our approach achieves state-of-the-art results and enables a robot to estimate the material class of household objects at almost 90\% accuracy, even when 92\% of the training data are unlabeled.

We show how a robot can classify an object's material with 90\% accuracy given half a second of contact and all labeled training samples. We compared the semi-supervised GAN to SVM and neural network classifiers trained on only labeled data, and we observed that learning from unlabeled data can improve performance considerably when few labeled examples are available. We also explore how well this generative adversarial approach can recognize the materials of new objects via leave-one-object-out cross-validation. GANs achieved $>$90\% classification accuracy when provided with a few labeled training samples for each object. However, a GAN's performance dropped by $\sim$20\% when classifying new objects with no labeled samples in the training set. It remains an open question how these generative models can more accurately recognize an object's material when only unlabeled samples from the object are available. Finally, to support further research, we have released the haptics dataset from this work, which was collected on a PR2 robot that performed 7,200 interactions with everyday household objects.

\clearpage
\acknowledgments{We would like to thank Joshua Wade and Tapomayukh Bhattacharjee for their assistance throughout this work, and the anonymous reviewers for providing feedback. We thank the developers of Keras, which we used throughout our experiments~\citep{chollet2015keras}.}


\bibliography{bibliography}  

\include{supplementary}

\end{document}

%% file: supplementary.tex




\appendix
\section{Supplementary Material}

\begin{figure}[!b]
\centering
\includegraphics[width=1.0\textwidth, trim={0cm 0cm 0cm 0cm}, clip]{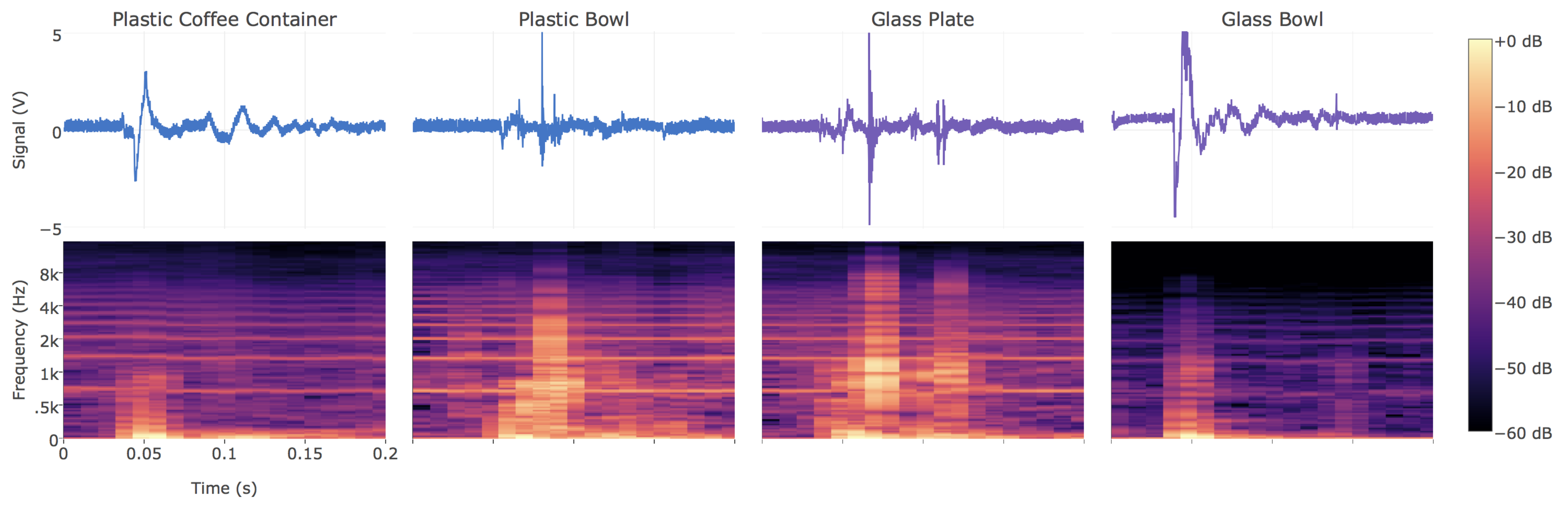}
\includegraphics[width=0.925\textwidth, left, trim={0cm 1cm 4cm 0cm}, clip]{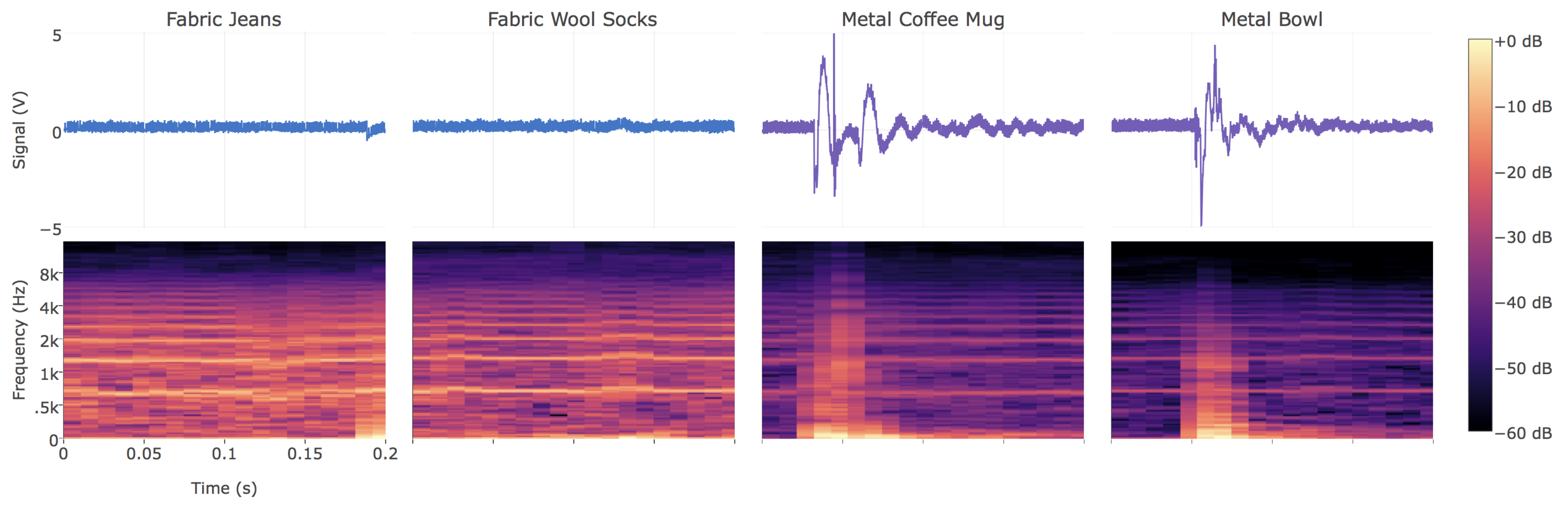}
\includegraphics[width=0.925\textwidth, left, trim={0cm 1cm 4cm 0cm}, clip]{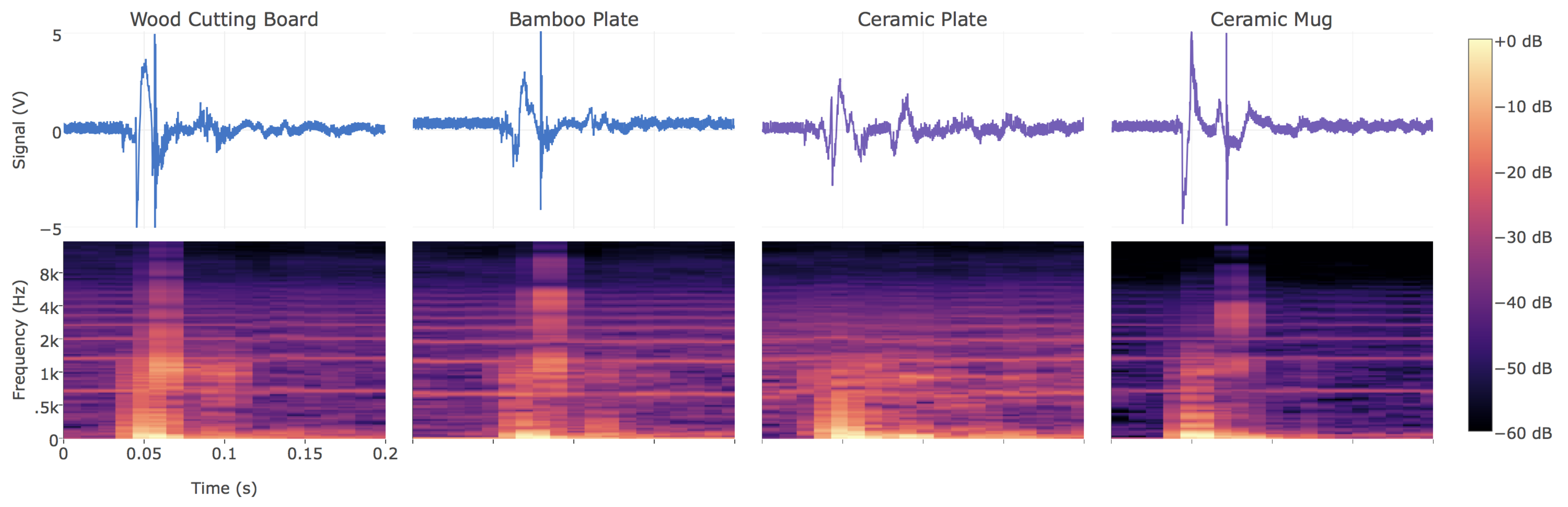}

\caption{\label{fig:spectrograms}Example contact microphone signals and associated Mel-scaled spectrograms for two objects from each material category.}
\end{figure}

We found that the vibration signals measured from a contact microphone appear valuable for rapid material classification. Specifically, when the robot made contact with an object for less than 0.5 seconds, these vibrations led to higher material recognition performance than either force or temperature sensing alone. Given these findings, we have prepared this section to provide more insight into these vibration signals. Figure~\ref{fig:spectrograms} displays vibration signals and associated Mel-scale spectrograms for two objects from each of the six material categories. The vibration signals are measured directly from the contact microphone, whereas the computed Mel-scale power spectrograms are provided as input features for learning.

In Figure~\ref{fig:spectrograms}, we see that fabrics are easily distinguishable from objects of other material types using vibration signals alone. In particular, these soft and compliant garments provide little discernible vibration information upon contact. Yet, note that these signals can be dependent on the interactions that a robot performs with an object. Although we had the robot perform rapid touching interactions, a long duration sliding interaction may provide more apparent vibrations for fabrics. On the other hand, plastic, glass, metal, wood, and ceramic all lead to clearer vibration signals upon impact between the robot's end effector and the object. We notice that contact with plastics tend to give less pronounced vibration signals, whereas metal objects typically produce more pronounced signals.

